\documentclass{article}

 \PassOptionsToPackage{numbers, compress}{natbib}

\usepackage{multicol,lipsum}
\usepackage[final]{nips_2018}
\usepackage{color}
\usepackage[utf8]{inputenc} 
\usepackage[T1]{fontenc}    
\usepackage{hyperref}       
\usepackage{url}            
\usepackage{booktabs}       
\usepackage{amsfonts}       
\usepackage{nicefrac}       
\usepackage{microtype}      
\usepackage{amsmath}
\usepackage{graphicx}
\usepackage{subcaption}
\usepackage{tikz}
\usepackage{amsmath}
\usepackage{color}
\usepackage{amsthm}

\makeatletter
\let\@fnsymbol\@arabic
\makeatother

\newtheorem{thm}{Theorem}[section]
\newtheorem{lem}[thm]{Lemma}

\title{How the Softmax Output is \\
Misleading for Evaluating \\
the Strength of Adversarial Examples} 

\author{
  Utku Ozbulak
  \thanks{Department of Electronics and Information Systems, Ghent University, Belgium} \,
  \thanks{Center for Biotech Data Science, Ghent University Global Campus, Republic of Korea}
   \qquad  Wesley De Neve
   \footnotemark[1] \,  \footnotemark[2] 
   \qquad Arnout Van Messem
   \footnotemark[2] \,
  \thanks{Department of Applied Mathematics, Computer Science and Statistics, Ghent University, Belgium}
  \\
  \texttt{\{utku.ozbulak,wesley.deneve,arnout.vanmessem\}@ugent.be}
}

\begin{document}
\maketitle
\begin{abstract}
Even before deep learning architectures became the de facto models for complex computer vision tasks, the softmax function was,  given its elegant properties, already used to analyze the predictions of feedforward neural networks. Nowadays, the output of the softmax function is also commonly used to assess the strength of adversarial examples: malicious data points designed to fail machine learning models during the testing phase. However, in this paper, we show that it is possible to generate adversarial examples that take advantage of some properties of the softmax function, leading to undesired outcomes when interpreting the strength of the adversarial examples at hand. Specifically, we argue that the output of the softmax function is a poor indicator when the strength of an adversarial example is analyzed and that this indicator can be easily tricked by already existing methods for adversarial example generation.
\end{abstract}






\section{Introduction}
Even though deep convolutional neural networks outperform other models on various computer vision problems such as image classification \cite{resnet, VGG}, object detection \citep{DBLP:journals/corr/RedmonDGF15}, and segmentation \citep{DBLP:journals/corr/RonnebergerFB15}, it has been shown that these models are not foolproof. A recent development called \textit{adversarial examples} currently stands as one of the major issues these models are facing \cite{Szegedy-Intriguingproperties}. Although there is no clear definition of an \textit{adversarial example}, we could call a sample adversarial if it is perturbed to be misclassified. As new and more optimized attack techniques are engineered on a regular basis, more complex defense mechanisms are proposed to counter adversarial examples. Nevertheless, assessing the viability of newly proposed defense mechanisms is not straightforward, attracting substantial criticism as they are not deemed sufficiently robust against \textit{strong} adversarial attacks \cite{athalye2018obfuscated, DBLP:journals/corr/CarliniW17}. However, no clear definition currently exists of what makes an adversarial example \textit{strong} or \textit{weak}.

As a consequence of its refined statistical properties, the softmax function is often used to analyze the prediction of a neural network \cite{bridle1990probabilistic, lecun1998gradient}. In this context, an adversarial example is usually referred to as \textit{strong} if it is predicted with high confidence by the model it is generated from and \textit{weak} if its confidence is low, where confidence is defined as the probabilistic outcome obtained by using the values of the logits (i.e., the raw output of the model) as input for the softmax function.



In this paper, we investigate the reliability of the softmax function for evaluating the strength of adversarial examples, or the lack thereof. We explain why the softmax function leads to poor judgment when it comes to identifying the strength of adversarial examples, and provide two concrete cases with examples from the ImageNet dataset \cite{ILSVRC15:rus}. Finally, we show our observations hold true across multiple models by presenting detailed experiments on the link between the output (e.g., softmax) and the transferability of adversarial examples for AlexNet, VGG-16, and ResNet-50 \cite{Alexnet, VGG, resnet}.


\section{Softmax for Adversarial Examples: Shortcomings and Consequences}
\label{Shortcomings}
When the prediction of a neural network is analyzed, the output is usually represented in terms of probabilities. To convert logits, hereafter also referred to as \textit{class activations} (CA), into probabilities, a normalized exponential function called the softmax function $P(\mathbf{u})_k = \dfrac{e^{\mathbf{u}_k}}{\sum_{m=1}^{M}e^{\mathbf{u}_m}}$ is used, where $\mathbf{u}$ is an input vector such that $\mathbf{u}=(u_1, \ldots, u_M)^T \in \mathbb{R}^{M}$ and  $k$ is the selected index of the vector $\mathbf{u}$~\citep{Bishop06a,Goodfellow-et-al-2016,bridle1990probabilistic}. In particular, the softmax function uses the exponential function to squeeze the input values between zero and one in such a way that the output values add up to one. This property makes the softmax function helpful in more easily interpreting the predictions of a neural network, instead of having to rely on the class activations, which are more difficult to interpret. The output of the softmax function is mostly referred to as the \textit{confidence} of the prediction made.


In reality, the softmax function has two drawbacks for correctly interpreting the predictions of a neural network when adversarial examples are at stake if it is used in settings with limited decimal precision. The first one is its lack of a unique input-to-output mapping (in other words, the function is not injective \cite{gao2017properties}); the second one is its sensitivity to high-magnitude inputs, which is due to its reliance on the exponential function. As we will show momentarily, these limitations can mask certain characteristics of adversarial examples, and could in some cases even be \textit{abused} by certain techniques for adversarial example generation.

\textbf{Over-optimized Adversarial Examples.} Multiple methods have been proposed to generate adversarial examples since their discovery by \citet{Szegedy-Intriguingproperties}. Most of these methods are based on an iterative approach for optimizing the adversarial examples \cite{DBLP:journals/corr/KurakinGB16, DBLP:journals/corr/CarliniW16a, Nguyen-deepnnseasilyfooled, Szegedy-Intriguingproperties, DBLP:journals/corr/PapernotMJFCS15}, making it possible to further optimize an adversarial example (in terms of the logit values), even after obtaining full confidence. However, once the prediction confidence is mapped to one, it is impossible to differentiate between the next iterations of the adversarial example based on the softmax output. Indeed, as the corresponding input (logit) keeps increasing, the softmax output will remain the same, as shown in Lemma~\ref{cor:1}.




\begin{figure}
\captionsetup[subfigure]{justification=centering}
    \centering
      \begin{subfigure}{0.22\textwidth}
        \includegraphics[width=\textwidth]{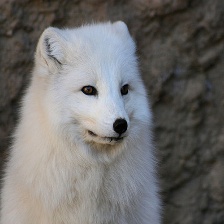}
          \caption{Original Image \\ Prediction: Arctic Fox \\Confidence: $0.99$ \\ $\text{CA}_1$: $\sim 20$ \\ $\text{CA}_2$: $\sim 5 \phantom{0}$}
      \end{subfigure}
      \begin{subfigure}{0.22\textwidth}
        \includegraphics[width=\textwidth]{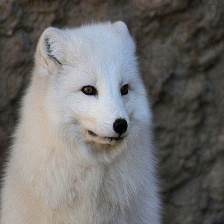}
          \caption{Adv. Image\\ Prediction: Radio \\Confidence: $1$ \\ $\text{CA}_1$: $\sim 1e2$ \\ $\text{CA}_2$: $\sim 10 \phantom{0}$}
      \end{subfigure}
      \begin{subfigure}{0.22\textwidth}
        \includegraphics[width=\textwidth]{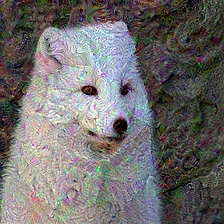}
          \caption{Adv. Image\\  Prediction: Radio \\Confidence: $1$ \\ $\text{CA}_1$: $\sim 5e2$ \\ $\text{CA}_2$: $\sim 12 \phantom{0}$}
      \end{subfigure}
      \begin{subfigure}{0.22\textwidth}
        \includegraphics[width=\textwidth]{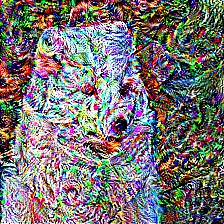}
          \caption{Adv. Image\\  Prediction: Radio \\Confidence: $1$ \\ $\text{CA}_1$: $\sim 4e3$ \\ $\text{CA}_2$: $\sim 40 \phantom{0}$}
      \end{subfigure}
      \caption{(a) Original image, predicted as \textit{arctic fox} with $0.99$ confidence. (b)-(c)-(d) Over-optimized adversarial examples, predicted by ResNet-50 with the same confidence, but with vastly different class activations. $\text{CA}_1$ and $\text{CA}_2$ are the class activations of the most and second most likely predictions,  respectively.}
      \label{fig:over-opt}
      \vspace{-1em}
\end{figure}

\begin{lem}\label{cor:1} 
When the softmax function is used in settings with limited decimal precision, it is no longer sensitive to positive changes in the magnitude of the largest input once the corresponding output has been mapped to one.
\end{lem}
Proof of Lemma~\ref{cor:1} can be found in Appendix A.

To show a practical outcome of Lemma~\ref{cor:1} for neural networks, we provide an original image in Figure~\ref{fig:over-opt}, classified as \textit{arctic fox} with $0.99$ confidence and three adversarial counterparts, all of which are classified as \textit{radio} with a confidence values of $1$ by a pretrained ResNet-50 \cite{resnet}. All of the three adversarial examples have been over-optimized to produce class activations that are beyond the reach of any natural image, with class activations approximately achieving values of $1e2$, $5e2$, and $4e3$, respectively. As a comparison, the highest class activation achieved by a genuine image in the whole Imagenet validation dataset~\cite{ILSVRC15:rus} for the pretrained ResNet-50 we use in this experiment is $\sim52$, and the highest activation for the target class $radio$ is only $\sim23$. As can be observed in Figure~\ref{fig:over-opt}, when an adversarial example is referred to as a \textit{high-confidence} adversarial example based on the output of the softmax function, both the amount of activation it produces and how far it is optimized are not clear, given the masking effect of the softmax function.


\textbf{Multi-class Optimized Adversarial Examples.}
In order to generate more robust adversarial examples, multi-class optimization was proposed by~\citet{DBLP:journals/corr/CarliniW16a}. This method aims to produce an adversarial example that is not only predicted as the targeted class with high confidence, but this method also optimizes the second most likely class so that the adversarial example can be easily transferred between models, almost surely being predicted as one of those two classes. In simple terms, this attack implicitly adds perturbations from two sources: the target class and the second most likely class. In this case, when the optimization is performed multiple times and the activations of these two targeted classes become much larger than all other class activations, then the confidence of the prediction will only depend on these two classes. This attack effectively takes advantage of how the softmax function maps inputs to outputs to disguise a \textit{strong} adversarial example as a \textit{weak} one. Using this attack, or any other multi-class optimization technique, it is therefore possible to generate adversarial examples that produce extremely high activations for the selected classes and that are still disguised as \textit{low-confidence} adversarial examples. 

\begin{lem}\label{lemma:2}
When the softmax function is used in settings with limited decimal precision, increasing the input which corresponds to the highest output after softmax function does not result in an increase in that output when the input of any other non-zero output has a larger increase.


\end{lem}
Proof of Lemma~\ref{lemma:2} can be found in Appendix A.





  \begin{figure}[t]
\captionsetup[subfigure]{justification=centering}
    \centering
      \begin{subfigure}{0.22\textwidth}
        \includegraphics[width=\textwidth]{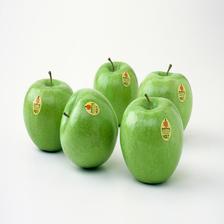}
          \caption{Original Image \\ Prediction: Apple\\Confidence: $0.95$ \\ $\text{CA}_1$: $\sim 19$  \\ $\text{CA}_2$: $ \sim 16$ \\ $\text{CA}_3$: $\sim 4 \phantom{0}$}
      \end{subfigure}
      \begin{subfigure}{0.22\textwidth}
        \includegraphics[width=\textwidth]{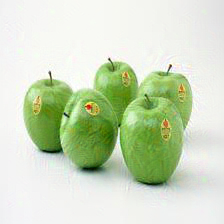}
          \caption{Adv. Image\\ Prediction: Radio \\Confidence: $0.75$ \\ $\text{CA}_1$: $\sim 39$  \\ $\text{CA}_2$: $\sim 38$\\  $\text{CA}_3$: $\sim 7 \phantom{0}$}
      \end{subfigure}
      \begin{subfigure}{0.22\textwidth}
        \includegraphics[width=\textwidth]{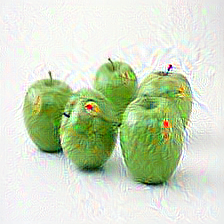}
          \caption{Adv. Image\\  Prediction: Radio \\Confidence: $0.71$ \\ $\text{CA}_1$: $\sim 190$  \\ $\text{CA}_2$: $\sim 189$ \\ $\text{CA}_3$: $\sim 10\phantom{0}$}
      \end{subfigure}
      \begin{subfigure}{0.22\textwidth}
        \includegraphics[width=\textwidth]{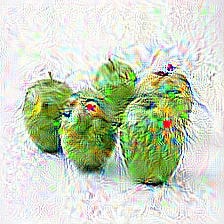}
          \caption{Adv. Image\\  Prediction: Radio \\Confidence: $0.57$ \\ $\text{CA}_1$: $\sim 523$  \\ $\text{CA}_2$: $\sim 522$  \\ $\text{CA}_3$: $\sim 14\phantom{0}$}
      \end{subfigure}
      \caption{(a) Original image, predicted as \textit{apple} with $0.95$ confidence. (b)-(c)-(d) Multi-class optimized adversarial examples that produce higher class activations, but that are predicted with lower confidence by ResNet-50. $\text{CA}_1$, $\text{CA}_2$, and $\text{CA}_3$ are the class activations of the first, second, and third most likely predictions, respectively.}
      \label{fig:multi-opt}
      \vspace{-1em}
\end{figure}
Practical examples for this lemma can be found in Figure~\ref{fig:multi-opt}, showing adversarial examples that are predicted with lower confidence than their predecessors, although the class activation of the corresponding input has increased. The prediction confidence of these adversarial examples almost entirely depends on two out of a thousand classes due to the vast difference among the magnitudes of the different predictions made. Naturally, since the softmax outputs only depend on two entries, the lowest confidence that can be achieved for this case is slightly higher than $0.50$. However, it is possible to extend this two-class attack to a larger multi-class attack, producing confidence values that are even lower, further disguising the adversarial examples when the softmax output is measured. This again shows that the output of the softmax function may give rise to misleading results when evaluating the strength of adversarial examples. Note that the results of Lemma~\ref{lemma:2} can easily be extended to larger multi-class schemes in which more than two classes are optimized. 





  \begin{figure}[t]
\captionsetup[subfigure]{justification=centering}
    \centering
      \begin{subfigure}{0.47\textwidth}
        \includegraphics[width=\textwidth]{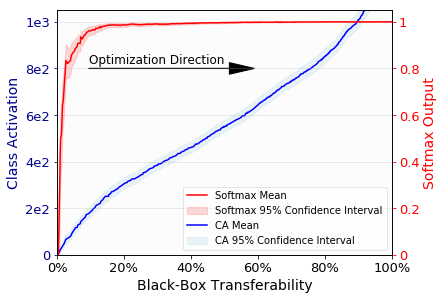}
          \caption{Over-optimized Adv. Examples}
      \end{subfigure}
      \begin{subfigure}{0.47\textwidth}
        \includegraphics[width=\textwidth]{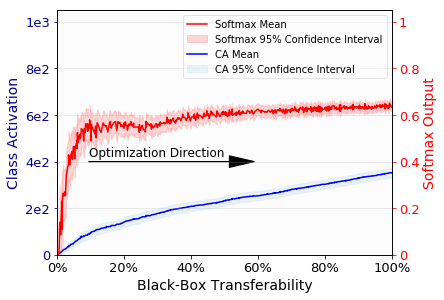}
          \caption{Multi-class Optimized Adv. Examples}
      \end{subfigure}
      \caption{Highest class activation and corresponding softmax output as a function of black-box transferability of adversarial examples. Adversarial examples are generated by VGG-16 and tested against ResNet-50.}
      \label{fig:line_graphs}
      \vspace{-1em}
\end{figure}

\textbf{Experimental Results on Transferability of Adversarial Examples.}
Although there is no perfect criterion to quantify the strength of an adversarial example, one way to measure it is to analyze whether or not the adversarial example at hand transfers to other models \cite{DBLP:journals/corr/CarliniW17}.

To demonstrate the practicality of our previous observations, we present two plots in Figure~\ref{fig:line_graphs}, detailing the black-box transferability of $2000$ adversarial examples, for both the case of (a) over-optimized and (b) multi-class optimized adversarial examples, with the genuine inceptions of the adversarial examples taken from the ImageNet validation dataset. Black-box transferability is measured over the course of adversarial optimization, showing its relation to the highest class activation and its related softmax output taken from the source model. In particular, the two plots in Figure~\ref{fig:line_graphs} show the mean and the corresponding 95\% confidence interval of the class activations and the softmax output of adversarial examples throughout the adversarial optimization as a function of black-box transferability. The adversarial examples are generated using VGG-16 \cite{VGG} and their transferability is tested against ResNet-50 \cite{resnet}.


As can be observed from Figure~\ref{fig:line_graphs}(a), the softmax confidence almost immediately jumps to $100\%$, making it from this point onwards impossible to differentiate between consecutive adversarial examples. Figure~\ref{fig:line_graphs}(b) shows that, in the case of multi-class optimization, the confidence is stuck at approximately $0.6$ throughout the graph, even though the transferability is increasing as the optimization is continued. Both plots in Figure~\ref{fig:line_graphs} show that the output of the softmax function is not a good indicator of the transferability (i.e., the \textit{strength}) of adversarial examples.




Detailed explanations of the experimental settings and the adversarial example generation methods used, as well as of further results covering multiple models, can be found in the Appendix B.
\section{Conclusions and Directions for Future Research}
In this paper, we showed that the softmax function is a poor indicator for determining the strength of adversarial examples. In support of this claim, we discussed two concrete cases: over-optimized and multi-class optimized adversarial examples. For these two cases, the outcome of the softmax function leads to misleading interpretations, whereas the class activations, in some cases, provide solid clues on the strength of the adversarial examples. As it stands, to correctly measure the strength of adversarial examples, a more reliable method is needed that is sensitive to both the magnitude and the distribution of the class activations.



Furthermore, all of the previously presented evidence does not only prevent the research community from correctly determining the strength of adversarial examples when they are used to assess a proposed defense technique, but the evidence also shows that the softmax function masks how easy it is to identify the certain types of adversarial examples with high class activations. To that end, strategically leveraging class activations as a first-line defense may already protect against a high number of adversarial attacks, even before triggering more complex defense mechanisms. 






\bibliographystyle{abbrvnat}
\bibliography{egbib}

\begin{thebibliography}{18}
\providecommand{\natexlab}[1]{#1}
\providecommand{\url}[1]{\texttt{#1}}
\expandafter\ifx\csname urlstyle\endcsname\relax
  \providecommand{\doi}[1]{doi: #1}\else
  \providecommand{\doi}{doi: \begingroup \urlstyle{rm}\Url}\fi

\bibitem[Athalye et~al.(2018)Athalye, Carlini, and
  Wagner]{athalye2018obfuscated}
A.~Athalye, N.~Carlini, and D.~Wagner.
\newblock Obfuscated gradients give a false sense of security: Circumventing
  defenses to adversarial examples.
\newblock \emph{arXiv preprint arXiv:1802.00420}, 2018.

\bibitem[Bishop(2006)]{Bishop06a}
C.~M. Bishop.
\newblock \emph{Pattern Recognition and Machine Learning}.
\newblock Springer, 2006.

\bibitem[Bridle(1990)]{bridle1990probabilistic}
J.~S. Bridle.
\newblock Probabilistic interpretation of feedforward classification network
  outputs, with relationships to statistical pattern recognition.
\newblock In \emph{Neurocomputing}, pages 227--236. Springer, 1990.

\bibitem[Carlini and Wagner(2016)]{DBLP:journals/corr/CarliniW16a}
N.~Carlini and D.~A. Wagner.
\newblock Towards evaluating the robustness of neural networks.
\newblock \emph{CoRR}, abs/1608.04644, 2016.

\bibitem[Carlini and Wagner(2017)]{DBLP:journals/corr/CarliniW17}
N.~Carlini and D.~A. Wagner.
\newblock Adversarial examples are not easily detected: Bypassing ten detection
  methods.
\newblock \emph{CoRR}, abs/1705.07263, 2017.

\bibitem[Gao and Pavel(2017)]{gao2017properties}
B.~Gao and L.~Pavel.
\newblock On the properties of the softmax function with application in game
  theory and reinforcement learning.
\newblock \emph{arXiv preprint arXiv:1704.00805}, 2017.

\bibitem[Goodfellow et~al.(2016)Goodfellow, Bengio, and
  Courville]{Goodfellow-et-al-2016}
I.~Goodfellow, Y.~Bengio, and A.~Courville.
\newblock \emph{Deep Learning}.
\newblock MIT Press, 2016.

\bibitem[He et~al.(2016)He, Zhang, Ren, and Sun]{resnet}
K.~He, X.~Zhang, S.~Ren, and J.~Sun.
\newblock Deep residual learning for image recognition.
\newblock In \emph{Proceedings of the IEEE conference on computer vision and
  pattern recognition}, pages 770--778, 2016.

\bibitem[Krizhevsky et~al.(2012)Krizhevsky, Sutskever, and Hinton]{Alexnet}
A.~Krizhevsky, I.~Sutskever, and G.~E. Hinton.
\newblock Imagenet classification with deep convolutional neural networks.
\newblock In F.~Pereira, C.~J.~C. Burges, L.~Bottou, and K.~Q. Weinberger,
  editors, \emph{Advances in Neural Information Processing Systems 25}, pages
  1097--1105. Curran Associates, Inc., 2012.

\bibitem[Kurakin et~al.(2016)Kurakin, Goodfellow, and
  Bengio]{DBLP:journals/corr/KurakinGB16}
A.~Kurakin, I.~Goodfellow, and S.~Bengio.
\newblock Adversarial examples in the physical world.
\newblock \emph{CoRR}, abs/1607.02533, 2016.

\bibitem[LeCun et~al.(1998)LeCun, Bottou, Bengio, and
  Haffner]{lecun1998gradient}
Y.~LeCun, L.~Bottou, Y.~Bengio, and P.~Haffner.
\newblock Gradient-based learning applied to document recognition.
\newblock \emph{Proceedings of the IEEE}, 86\penalty0 (11):\penalty0
  2278--2324, 1998.

\bibitem[Nguyen et~al.(2015)Nguyen, Yosinski, and
  Clune]{Nguyen-deepnnseasilyfooled}
A.~Nguyen, J.~Yosinski, and J.~Clune.
\newblock Deep neural networks are easily fooled: High confidence predictions
  for unrecognizable images.
\newblock In \emph{Proceedings of the IEEE Conference on Computer Vision and
  Pattern Recognition}, pages 427--436, 2015.

\bibitem[Papernot et~al.(2015)Papernot, McDaniel, Jha, Fredrikson, Celik, and
  Swami]{DBLP:journals/corr/PapernotMJFCS15}
N.~Papernot, P.~D. McDaniel, S.~Jha, M.~Fredrikson, Z.~B. Celik, and A.~Swami.
\newblock The limitations of deep learning in adversarial settings.
\newblock \emph{CoRR}, abs/1511.07528, 2015.

\bibitem[Redmon et~al.(2015)Redmon, Divvala, Girshick, and
  Farhadi]{DBLP:journals/corr/RedmonDGF15}
J.~Redmon, S.~K. Divvala, R.~B. Girshick, and A.~Farhadi.
\newblock You only look once: Unified, real-time object detection.
\newblock \emph{CoRR}, abs/1506.02640, 2015.

\bibitem[Ronneberger et~al.(2015)Ronneberger, Fischer, and
  Brox]{DBLP:journals/corr/RonnebergerFB15}
O.~Ronneberger, P.~Fischer, and T.~Brox.
\newblock U-net: Convolutional networks for biomedical image segmentation.
\newblock In N.~Navab, J.~Hornegger, W.~M. Wells, and A.~F. Frangi, editors,
  \emph{Medical Image Computing and Computer-Assisted Intervention -- MICCAI
  2015}, pages 234--241, Cham, 2015. Springer International Publishing.
\newblock ISBN 978-3-319-24574-4.

\bibitem[Russakovsky et~al.(2015)Russakovsky, Deng, Su, Krause, Satheesh, Ma,
  Huang, Karpathy, Khosla, Bernstein, Berg, and Fei-Fei]{ILSVRC15:rus}
O.~Russakovsky, J.~Deng, H.~Su, J.~Krause, S.~Satheesh, S.~Ma, Z.~Huang,
  A.~Karpathy, A.~Khosla, M.~Bernstein, A.~C. Berg, and L.~Fei-Fei.
\newblock {ImageNet Large Scale Visual Recognition Challenge}.
\newblock \emph{International Journal of Computer Vision}, 115\penalty0
  (3):\penalty0 211--252, 2015.

\bibitem[Simonyan and Zisserman(2014)]{VGG}
K.~Simonyan and A.~Zisserman.
\newblock Very deep convolutional networks for large-scale image recognition.
\newblock \emph{arXiv preprint arXiv:1409.1556}, 2014.

\bibitem[Szegedy et~al.(2013)Szegedy, Zaremba, Sutskever, Bruna, Erhan,
  Goodfellow, and Fergus]{Szegedy-Intriguingproperties}
C.~Szegedy, W.~Zaremba, I.~Sutskever, J.~Bruna, D.~Erhan, I.~Goodfellow, and
  R.~Fergus.
\newblock Intriguing properties of neural networks.
\newblock \emph{CoRR}, abs/1312.6199, 2013.

\end{thebibliography}

\appendix
\section*{\huge{Appendix A}}
\label{appendixA}
This appendix contains the omitted proofs in the main document.

\textbf{Lemma 2.1.}\textit{
When the softmax function is used in settings with limited decimal precision, it is no longer sensitive to positive changes in the magnitude of the largest input once the corresponding output has been mapped to one.
}

\begin{proof}
Take $\mathbf{u}=(u_1, u_2, \ldots, u_M)^T \in \mathbb{R}^{M}$ with $u_1 \in \mathbb{R}^+$ such that $P(\mathbf{u}) = (p_1, p_2, \ldots, p_M)^T$, $p_1>p_i \,, \forall \ i \in \{2,\ldots,M\}$ and assume that the calculation is performed under a limited decimal precision where the smallest positive number that can be represented is $\delta$, $\, \delta \in (0, 0.1)$. Under these conditions, if $p_1 > 1-\delta$, then $p_1 \equiv 1$, and  $\forall v_1 \in \mathbb{R}^+$ with $v_1>u_1$, $P(v_1, u_2, \ldots, u_M)^T = (1, 0, \ldots, 0)^T$.
\end{proof}

\textbf{Lemma 2.2.}\textit{
When the softmax function is used in settings with limited decimal precision, increasing the input which corresponds to the highest output after softmax function does not result in an increase in that output when the input of any other non-zero output has a larger increase.
}
\begin{proof}
Take $\mathbf{u} \in \mathbb{R}^{M}$ such that $P(u_1, u_2, \ldots, u_M)^T=(p_1, p_2, \ldots, p_M)^T$, with $p_1>p_i$, $\forall \ i \in \{2,\ldots,M\}$ and assume that the calculation is performed under a limited decimal precision where the smallest positive number that can be represented is $\delta$, $\,\delta \in (0, 0.1)$. If $p_n < \delta$, $n \in \{1, 2, \ldots, M\}$, then $p_n \equiv 0$, and if $p_n > 1-\delta$, then $p_n \equiv 1$. For any $t \in \{2,\ldots,M\}$ with $p_t>0$, take $\mathbf{v} \in \mathbb{R}^{M}$ such that $v_t>v_1>0$, $u_1+v_1 > u_t + v_t$, and $v_k=0$ for $k \notin \{1,t\}$, then $P(\mathbf{u} + \mathbf{v})^T = (l_1, l_2, \ldots, l_M)^T$, with $l_1 > l_t$, but $l_1 < p_1$.
\end{proof}

\section*{\huge{Appendix B}}
\label{appendixB}
This appendix briefly discusses the methods used for adversarial example generation, as well as a number of additional experimental results for measuring black-box transferability of adversarial examples, and its relation with highest class activation and softmax output.

\section*{Notation and Framework}
\subsection*{Notation}
\begin{itemize}
\item $\mathbf{X}$ \textemdash \, an image represented as a 3-D tensor (depth $\times$ width $\times$ height) with integer values in the range $[0, 255]$.
\item $y = g(\theta, \mathbf{X})$ \textemdash \, a classification function that links $\mathbf{X}$ to a target label, using a neural network with parameters $\theta$. This neural network does not have a final softmax layer.
\item $\nabla_{x} g(\theta, \mathbf{X})_c$ \textemdash \, the partial derivative of a neural network $g$ with respect to $\mathbf{X}$ for a target label $c$.
\end{itemize}

\subsection*{Framework}

For the experiments on the transferability of adversarial examples, we used three well-known deep learning architectures: AlexNet, VGG-16, and ResNet-50 \cite{Alexnet, VGG, resnet}. To show that our observations hold true for a variety of models, we investigated all possible combinations of the three models.


\section*{Adversarial Example Generation Methods}
In order to be able to generate large number of adversarial examples quickly, we aimed to minimize the computational cost as much as possible. To achieve this, we used the simplest way to generate adversarial examples, not paying attention to the perturbation visibility or the $L_0$, $L_2$, or $L_\infty$ distances between the original image and the perturbed one. We would like to make note that when the distance loss is incorporated into the optimization, the results in terms of class activation and sofmax output will not change, but depending on the selected distance metric, the perturbation is added in a more subtle way at the cost of more computational complexity. The only constraint we set on our adversarial example generation method was the discretization constraint, which ensured that the generated adversarial examples were valid images: $\mathbf{X} \in [0, 255]^{n}$.


\subsection*{Generating Over-optimized Adversarial Examples}
We generated over-optimized adversarial examples using the following equation for the optimization: $\mathbf{X}_{i+1} = \mathbf{X}_i - \alpha \nabla_{x} g(\theta, \mathbf{X}_i)_c$. This attack is similar to proposed iterative attacks such as the basic iterative method \cite{DBLP:journals/corr/KurakinGB16} or box-constrained L-BFGS \cite{Szegedy-Intriguingproperties} in the sense that it aims to increase the target likelihood of class $c$ with each iteration. However, instead of using the cross entropy loss, we directly used the gradients from the model for the sake of speed. For the perturbation multiplier, we used $\alpha= 0.15$ to make sure the perturbation itself is reasonable and not extremely large.

\subsection*{Generating  Multi-class Optimized Adversarial Examples} 
To generate multi-class optimized adversarial examples, the following equation was used: $\mathbf{X}_{i+1} = \mathbf{X}_i - \alpha \nabla_{x} g(\theta, \mathbf{X}_i)_c - \beta \nabla_{x} g(\theta, \mathbf{X}_i)_d$. This optimization adds perturbations from two sources, namely the target classes $c$ and $d$, in order to simultaneously maximize the prediction likelihood of both classes. This procedure is similar to the Carlini-Wagner Attack \cite{DBLP:journals/corr/CarliniW16a}, however, instead of holding the difference between likelihood of both optimized classes $c$ and $d$ constant with a complex constraint (i.e., the parameter $k$ for the loss function in the original paper), we dynamically adjusted the perturbation multipliers. In order to prevent the prediction from alternating rapidly between both classes, we lowered the perturbation multipliers as compared to the previous method. We chose $\alpha, \beta \in [0.05, 0.1]$ and made sure that the multiplier of the less likely prediction class was larger than the other one. This approach ensured that the prediction did not reach full confidence in terms of softmax output and always stayed within the range of $0.5$ and $1$ for either one class or the other.

  \begin{figure}[t]
\captionsetup[subfigure]{justification=centering}
    \centering
      \begin{subfigure}{0.47\textwidth}
        \includegraphics[width=\textwidth]{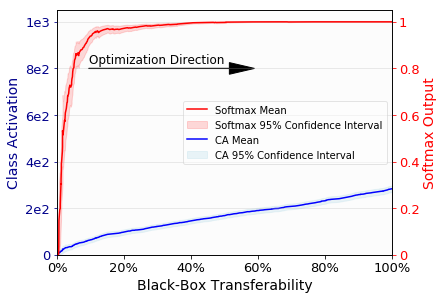}
          \caption{AlexNet to VGG-16}
      \end{subfigure}
      \begin{subfigure}{0.47\textwidth}
        \includegraphics[width=\textwidth]{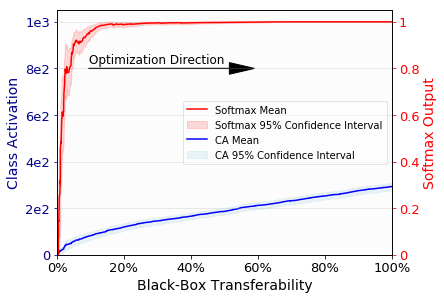}
          \caption{AlexNet to ResNet-50}
      \end{subfigure}
            \begin{subfigure}{0.47\textwidth}
        \includegraphics[width=\textwidth]{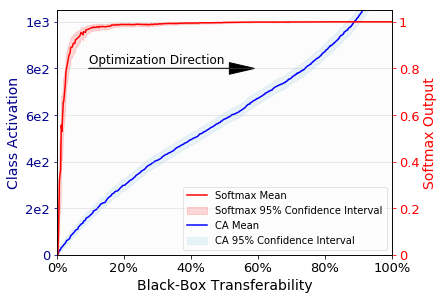}
          \caption{VGG-16 to AlexNet}
      \end{subfigure}
            \begin{subfigure}{0.47\textwidth}
        \includegraphics[width=\textwidth]{graphs/raw_vgg_to_res.png}
          \caption{VGG-16 to ResNet-50}
      \end{subfigure}
            \begin{subfigure}{0.47\textwidth}
        \includegraphics[width=\textwidth]{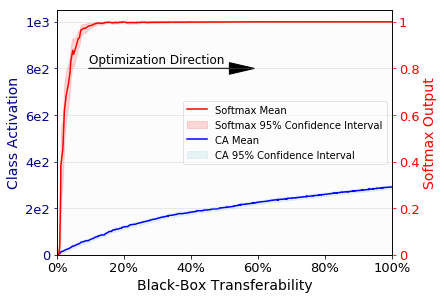}
          \caption{ResNet-50 to VGG-16}
      \end{subfigure}
        \begin{subfigure}{0.47\textwidth}
        \includegraphics[width=\textwidth]{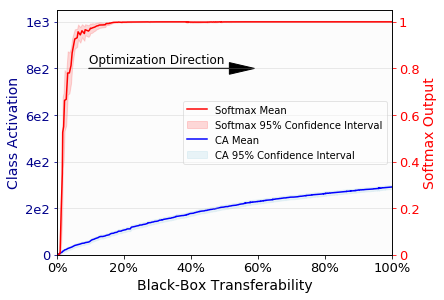}
          \caption{ResNet-50 to AlexNet}
      \end{subfigure}
      \caption{Highest class activation and corresponding softmax output as a function of black-box transferability of over-optimized adversarial examples. Adversarial examples are generated by AlexNet, VGG-16, or ResNet-50 and tested against each other.}
      \label{fig:over-opt-charts}
      \end{figure}

\section*{Experimental Results on the Transferability of Adversarial Examples}
\subsection*{Experimental Settings}
We first used $2500$ images of the ImageNet validation set to generate adversarial examples, making sure that a selected image is initially correctly predicted by all three models. Using the optimization techniques presented above, for the two types of adversarial examples under consideration (i.e., over-optimized and multi-class optimized), we generated $2500$ adversarial examples per model, totaling up to $7500$ adversarial examples for each type of adversarial example. These adversarial examples are optimized for over $500$ iterations; after each iteration for a particular adversarial example, we saved the highest class activation, its corresponding softmax output, and the transferability of that adversarial example against the other two models. We then processed the collected data using the following steps:

\textbf{Data Cleaning.} 
One might think that with enough perturbation an image eventually becomes adversarial after many iterations. This is, however, not always the case. For certain image-target class combinations, the optimization may halt, meaning that it may have gotten stuck in a local minimum, thus making it impossible to further alter the image. To remove these cases from our results, we filtered out all adversarial examples that did not transfer after $500$ iterations. This filtering operation removed $200$ to $500$ adversarial examples from the individual model-to-model adversarial example pool, leaving the results obtained for approximately $2000$ adversarial examples to work with. 


  \begin{figure}[t]
      \captionsetup[subfigure]{justification=centering}
    \centering
      \begin{subfigure}{0.47\textwidth}
        \includegraphics[width=\textwidth]{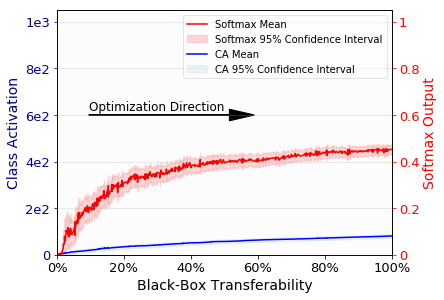}
          \caption{AlexNet to VGG-16}
      \end{subfigure}
      \begin{subfigure}{0.47\textwidth}
        \includegraphics[width=\textwidth]{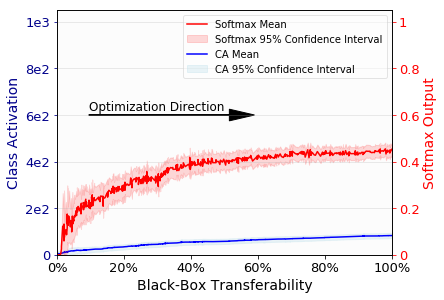}
          \caption{AlexNet to ResNet-50}
      \end{subfigure}
            \begin{subfigure}{0.47\textwidth}
        \includegraphics[width=\textwidth]{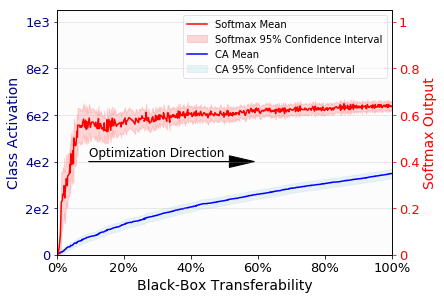}
          \caption{VGG-16 to AlexNet}
      \end{subfigure}
      \begin{subfigure}{0.47\textwidth}
        \includegraphics[width=\textwidth]{graphs/multi_vgg_to_res.png}
          \caption{VGG-16 to ResNet-50}
      \end{subfigure}
            \begin{subfigure}{0.47\textwidth}
        \includegraphics[width=\textwidth]{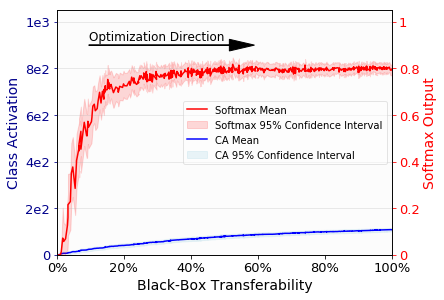}
          \caption{ResNet-50 to AlexNet}
      \end{subfigure}
            \begin{subfigure}{0.47\textwidth}
        \includegraphics[width=\textwidth]{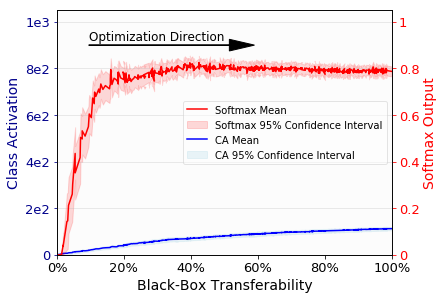}
          \caption{ResNet-50 to VGG-16}
      \end{subfigure}
      \caption{Highest class activation and corresponding softmax output as a function of black-box transferability of multi-class optimized adversarial examples. Adversarial examples are generated by AlexNet, VGG-16, or ResNet-50 and tested against each other.}
      \label{fig:multi-opt-charts}
\end{figure}

 \textbf{Data Aggregation.} 
 We calculated means and confidence intervals for every optimization step based on the data retained after the previously explained data cleaning operation. Each step of the optimization corresponds to a point in the $X$-axis which is represented as percentage of transferability (between $0\%$ and $100\%$) instead of optimization step (between $1$ and $500$). One might assume that a subsequent iteration might have less transferability than a previous one (i.e., the $(n+1)$th step has less transferability than the $n$th step), impairing the logic for this way of aggregation. However, we observed that, when an adversarial example is transferred to another model, more often than not, in the subsequent steps of the optimization, it continued to transfer. Thus, filtering the results based on transferability provided a decent $X$-axis, since the transferability started as low as $0\%$ (meaning that no adversarial example was able to transfer at the beginning) and eventually reached $100\%$ (meaning that all adversarial examples transferred in the end).

\subsection*{Experimental Results}
Figure~\ref{fig:over-opt-charts} and Figure~\ref{fig:multi-opt-charts} show the results obtained for over-optimized adversarial examples and multi-class optimized adversarial examples, respectively, using the previously explained experimental conditions for AlexNet, VGG-16, and ResNet-50. Given these plots, we can observe the following results:

\begin{itemize}
\item An adversarial example quickly reaches the state of being over-optimized, even with basic gradient ascent optimization, making the adversarial example generated during the next step already impossible to differentiate from the previous one when looking at the softmax output.

\item It is surprisingly easy to mask a \textit{strong} adversarial example as a \textit{weak} one based on the output of the softmax function. The results from Figure~\ref{fig:multi-opt-charts} show that, in the end, almost all of the adversarial examples are predicted with less than $100\%$ confidence by the generated model for the multi-class case, although all of them transferred.

\item Compared to the output of the softmax function, class activations provide better clues on the strength of adversarial examples. However, only relying on class activations is also not sufficient. Indeed, as for instance shown in Figure~\ref{fig:multi-opt-charts}, the average magnitude of class activations diminishes quite fast when the optimization targets multiple classes. 

\item Even though there are differences between the results obtained for each model, the overall behavior of the softmax output with respect to adversarial transferability is the same for all three models, showing that our observations hold true, regardless of the selected architecture.

\end{itemize}

\end{document}